\documentclass[letterpaper, 10 pt, conference]{format/ieeeconf}
\IEEEoverridecommandlockouts
\usepackage{multicol}
\usepackage[bookmarks=true]{hyperref}
\usepackage{amsmath,amsfonts,amssymb,mathrsfs}
\usepackage{verbatim}
\usepackage[vlined,ruled,linesnumbered]{algorithm2e}
\usepackage{wrapfig}
\usepackage{graphicx} 
\usepackage{xspace}

\DeclareMathOperator*{\argmin}{argmin}

\newcommand*{\qed}{\hfill\ensuremath{\square}}%

\title{\LARGE \bf Fast, Anytime Motion Planning for Prehensile
  Manipulation in Clutter}
\author{Andrew Kimmel, Rahul Shome, Zakary Littlefield, Kostas Bekris
\thanks{The authors are with the Computer Science Dept. of
  Rutgers University, 110 Frelinghuysen Road,
  Piscataway, NJ, USA, {\tt\small \{andrew.kimmel, rahul.shome, zakary.littlefield, kostas.bekris\}@rutgers.edu}.}}

\begin{document}

\maketitle
\thispagestyle{empty}
\pagestyle{empty}

\setlength{\abovecaptionskip}{-2.5pt}
\setlength{\belowcaptionskip}{-6pt}
\setlength{\dbltextfloatsep}{2pt plus 1.0pt minus 1.0pt}
\setlength{\textfloatsep}{2pt plus 1.0pt minus 1.0pt}
\setlength{\intextsep}{2pt plus 1.0pt minus 1.0pt}
\setlength{\belowdisplayskip}{0pt} \setlength{\belowdisplayshortskip}{2pt}
\setlength{\abovedisplayskip}{0pt} \setlength{\abovedisplayshortskip}{2pt}

\newcommand{\mam}{$\mathcal{G}_{\tt MAM}$}
\newcommand{\pr}{\ensuremath{\mathbb{P}}}

\newcommand{\reals}{\mathbb{R}}
\newcommand{\integers}{\mathbb{Z}}

\newcommand{\Wspace}{\mathbb{W}}
\newcommand{\Sspace}{\mathscr{S}}
\newcommand{\Robots}{R}
\newcommand{\Manip}{\mathbb{A}}
\newcommand{\Obstacles}{\mathscr{Z}}
\newcommand{\obst}{\mathbb{Z}}
\newcommand{\cspace}{\ensuremath{\mathbb{C}}}
\newcommand{\tspace}{\ensuremath{\mathbb{Q}}}
\newcommand{\constraints}{\ensuremath{\mathbb{M}}}
\newcommand{\ccross}{\ensuremath{\mathbb{C}}}
\newcommand{\ccrossfree}{\ensuremath{\mathbb{C}^{\textup{f}}}}
\newcommand{\cinv}{\cspace^{\textup{o}}}
\newcommand{\cbound}{\cspace_{\cap}}
\newcommand{\cstable}{\cspace_{s}}
\newcommand{\cgrasp}{\cspace_{G}}

\newcommand{\oracle}{\mathbb{O}_d}

\newcommand{\qrand}{\ensuremath{q_{\textup{rand}}}}
\newcommand{\vnear}{\ensuremath{V^{\textup{near}}}}
\newcommand{\vnew}{\ensuremath{V^{\textup{new}}}}
\newcommand{\vlast}{\ensuremath{V^{\textup{last}}}}
\newcommand{\vparent}{\ensuremath{V^{\textup{best}}}}

\newcommand{\Pspace}{\mathbb{P}}
\newcommand{\pose}{p}

\newcommand{\Qspace}{\mathbb{Q}}
\newcommand{\GeomManip}{\mathbb{WM}}

\newcommand{\rad}{\ensuremath{r(n)}}
\newcommand{\radstar}{\ensuremath{r^*(n)}}
\newcommand{\radi}{\ensuremath{r_i(n)}}
\newcommand{\radj}{\ensuremath{r_j(n)}}
\newcommand{\crossrad}{\ensuremath{r_R(n)}}
\newcommand{\crossradstar}{\ensuremath{r^*_R(n)}}
\newcommand{\impcrossrad}{\ensuremath{\hat r_R(n)}}
\newcommand{\allimpcrossrad}{\ensuremath{\hat r_{R}(n^R)}}
\newcommand{\ki}{\ensuremath{k_i(n)}}
\newcommand{\kj}{\ensuremath{k_j(n)}}

\newcommand{\mmgraph}{\ensuremath{\mathbb{G}}}
\newcommand{\mmgimp}{\hat\mmgraph}
\newcommand{\mmgexp}{\mmgraph}
\newcommand{\graph}{\ensuremath{\mathbb{G}}}
\newcommand{\aograph}{\ensuremath{\mathbb{G}^{AO}}}
\newcommand{\tree}{\ensuremath{\mathbb{T}}}
\newcommand{\mmnodes}{\mathbb{\hat V}}
\newcommand{\mmedges}{\mathbb{\hat E}}
\newcommand{\mmnodestpprm}{\mathbb{V}_{\chi_i}}
\newcommand{\mmedgestpprm}{\mathbb{E}_{\chi_i}}
\newcommand{\mmnode}{\mathbb{\hat v}}
\newcommand{\mmedge}{\mathbb{\hat e}}
\newcommand{\nodes}{\mathbb{V}}
\newcommand{\node}{\mathbb{v}}
\newcommand{\edges}{\mathbb{E}}
\newcommand{\edge}{\mathbb{e}}
\newcommand{\prmstar}{\ensuremath{ {\tt PRM^*} }}
\newcommand{\sprmstar}{Soft-\ensuremath{ {\tt PRM} }}
\newcommand{\irs}{\ensuremath{ {\tt IRS} }}
\newcommand{\spars}{{\tt SPARS}}
\newcommand{\drrt}{\ensuremath{{\tt dRRT}}}
\newcommand{\drrtstar}{\ensuremath{{\tt dRRT^*}}}
\newcommand{\dadrrtstar}{\ensuremath{{\tt dadRRT^*}}}

\newcommand{\sig}{{\tt SIG}}
\newcommand{\local}{\mathbb{L}}
\newcommand{\rmaps}{\ensuremath{\mathfrak{R}}}

\newcommand{\prm}{{\tt PRM}}
\newcommand{\mmprm}{\ensuremath{\text{Random-}{\tt MMP}}}
\newcommand{\kprmstar}{{\tt k-PRM$^*$}}
\newcommand{\rrt}{\ensuremath{{\tt RRT}}}
\newcommand{\rrtdrain}{{\tt RRT-Drain}}
\newcommand{\rrg}{{\tt RRG}}
\newcommand{\est}{{\tt EST}}
\newcommand{\rrtstar}{\ensuremath{\tt RRT^{\text *}}}
\newcommand{\astar}{{\ensuremath{\tt A^{\text *}}}}
\newcommand{\mstar}{{\tt M^{\text *}}}
\newcommand{\opens}{P_{Heap}}

\newcommand{\bvp}{{\tt BVP}}
\newcommand{\alg}{{\tt ALG}}
\newcommand{\fixed}{{\tt Fixed}-$\alpha$-\rdg}

\newcommand{\config}{C}

\newcommand{\cost}{\textup{cost}}

\newenvironment{myitem}{\begin{list}{$\bullet$}
{\setlength{\itemsep}{-0pt}
\setlength{\topsep}{0pt}
\setlength{\labelwidth}{0pt}
\setlength{\leftmargin}{10pt}
\setlength{\parsep}{-0pt}
\setlength{\itemsep}{0pt}
\setlength{\partopsep}{0pt}}}%
{\end{list}}

\newtheorem{claim}{\bf Claim}

\newcommand{\kiril}[1]{{\color{blue} \textbf{Kiril:} #1}}
\newcommand{\chups}[1]{{\color{red} \textbf{Chuples:} #1}}
\newcommand{\rahul}[1]{{\color{green} \textbf{Rahul:} #1}}

\newcommand{\T}{\mathcal{T}}

\newcommand{\leftrm}{\ensuremath{\mathbb{R}_{l}}  }
\newcommand{\rightrm}{\ensuremath{\mathbb{R}_{r}}  }
\newcommand{\leftmetric}{\ensuremath{\mathbb{P}_{l}}  }
\newcommand{\rightmetric}{\ensuremath{\mathbb{P}_{r}}  }
\newcommand{\cfull}{\ensuremath{\mathbb{C}_{{full}}}  }
\newcommand{\ceval}{\ensuremath{\mathbb{C}_{{eval}}}  }
\newcommand{\cfree}{\ensuremath{\mathbb{C}_{{free}}}  }
\newcommand{\cobs}{\ensuremath{\mathbb{C}_{{obs}}}  }
\newcommand{\cleft}{\ensuremath{\mathbb{C}_{{left}}}  }
\newcommand{\cright}{\ensuremath{\mathbb{C}_{{right}}}  }
\newcommand{\cshared}{\ensuremath{\mathbb{C}_{{shared}}}  }
\newcommand{\cgoal}{\ensuremath{q_{{goal}}}  }
\newcommand{\cstart}{\ensuremath{q_{{start}}}  }
\newcommand{\cend}{\ensuremath{q_{{end}}}  }

\newcommand{\gimpleft}{\ensuremath{\hat\mmgraph_l}}
\newcommand{\gimpright}{\ensuremath{\hat\mmgraph_r}}

\newcommand{\xrand}{\ensuremath{x^{\textup{rand}}}}
\newcommand{\xnear}{\ensuremath{x^{\textup{near}}}}
\newcommand{\xnew}{\ensuremath{x^{\textup{new}}}}
\newcommand{\xlast}{\ensuremath{x^{\textup{last}}}}
\newcommand{\xparent}{\ensuremath{x^{\textup{best}}}}

\newcommand{\sethree}{\ensuremath{\mathbb{SE}(3)}}
\newcommand{\obj}{\textit{o}}
\newcommand{\ctransit} {\ensuremath {\mathbb{C}_{transit}}}
\newcommand{\cmove}{\ensuremath{\mathbb{C}_{transfer}}}
\newcommand{\pregrasp}{\ensuremath{\mathbb{G}_{pre}}}
\newcommand{\pregraspset}{\ensuremath{\hat{\mathbb{G}}_{pre}}}
\newcommand{\cpre}{\ensuremath{q_{pre}}}
\newcommand{\epre}{\ensuremath{x_{pre}}}
\newcommand{\fk}{\ensuremath{FK}}
\newcommand{\ik}{\ensuremath{IK}}
\newcommand{\isst}{{\tt{iSST}}}
\newcommand{\jist}{{\tt{JIST}}}
\newcommand{\chomp}{{\tt{CHOMP}}}
\newcommand{\hval}{\ensuremath{h}}
\newcommand{\acand}{\ensuremath{\mathbb{A}_{cand}}}
\newcommand{\band}{ \mathbf{and}}
\newcommand{\bor}{ \mathbf{or}}
\newcommand{\eeroadmap}{\ensuremath{R_{ee}}}
\newcommand{\cdist}{\ensuremath{\mathtt{DISP}}}
\newcommand{\jacsteering}{\ensuremath{ \mathbb{J}^+\mathtt{steering}}}
\newcommand{\pseudoj}{\ensuremath{ \mathbb{J}^+}}

\newtheorem{definition}{Definition}
\newtheorem{assumption}{Assumption}

\begin{abstract}
  Many methods have been developed for planning the motion of robotic arms for picking and placing, ranging from local optimization to global search techniques, which are effective for sparsely placed objects.  Dense clutter, however, still adversely affects the success rate, computation times, and quality of solutions in many real-world setups. The current work integrates tools from existing methodologies and proposes a framework that achieves high success ratio in clutter with anytime performance. The idea is to first explore the lower dimensional end effector's task space efficiently by ignoring the arm, and build a discrete approximation of a navigation function, which guides the end effector towards the set of available grasps or object placements. This is performed online, without prior knowledge of the scene. Then, an informed sampling-based planner for the entire arm uses Jacobian-based steering to reach promising end effector poses given the task space guidance.  While informed, the method is also comprehensive and allows the exploration of alternative paths over time if the task space guidance does not lead to a solution.  This paper evaluates the proposed method against alternatives in picking or placing tasks among varying amounts of clutter for a variety of robotic manipulators with different end-effectors. The results suggest that the method reliably provides higher quality solution paths quicker, with a higher success rate relative to alternatives. 
\end{abstract}

\section{Introduction}
\label{sec:intro}

\begin{figure}[t]
\centering
\includegraphics[width=0.232\textwidth]{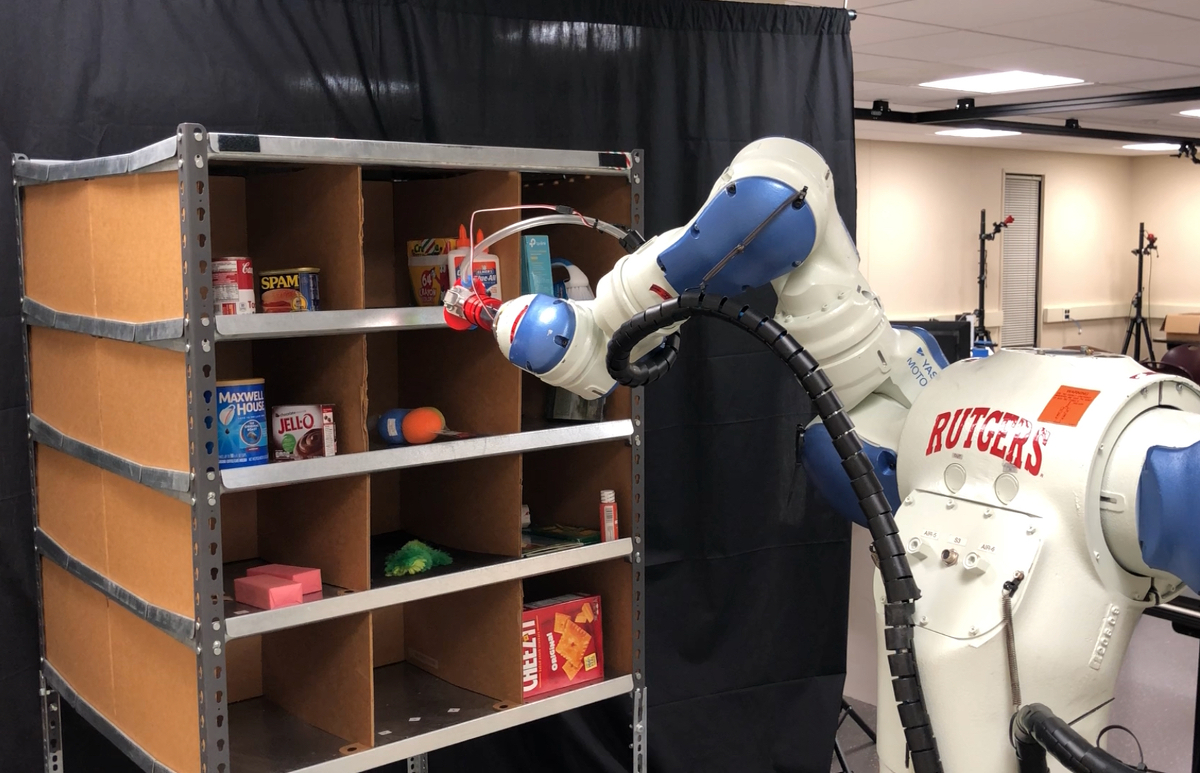}
\includegraphics[width=0.232\textwidth]{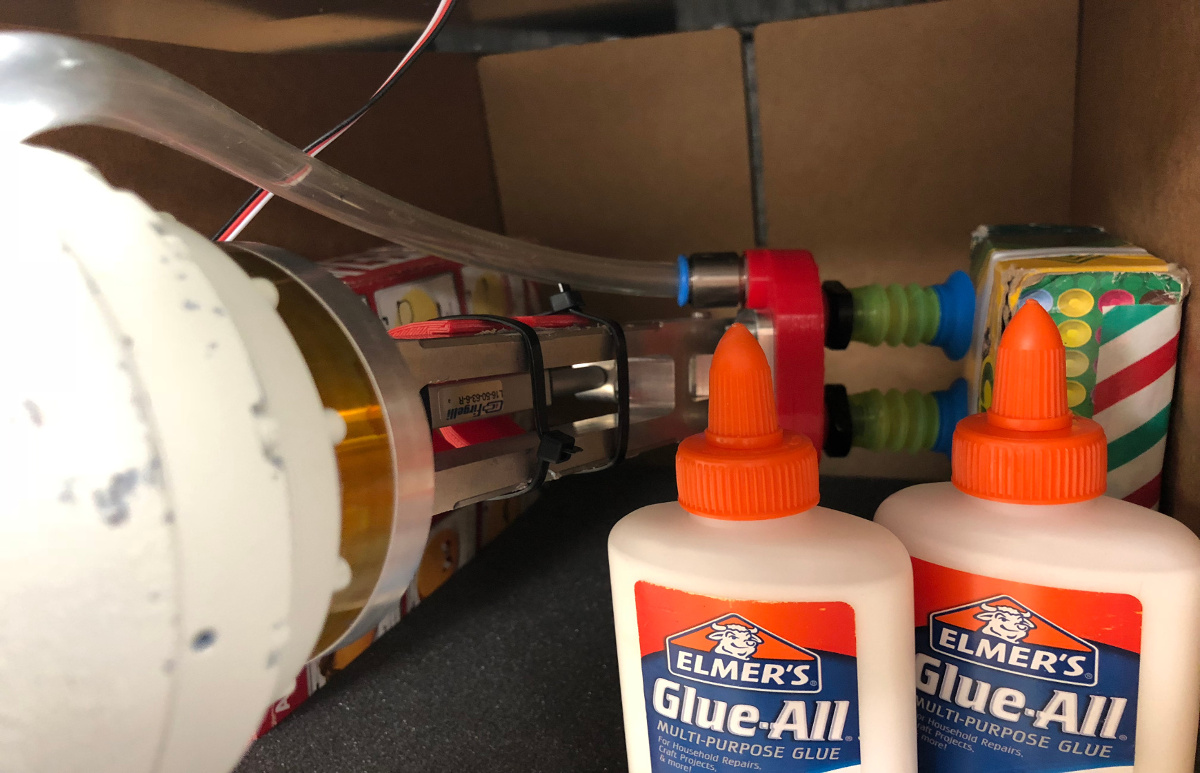}
\caption{An example of manipulation in cluttered table and shelf environments,
where the grasped object is partially occluded by other objects.}
\label{fig:table_clutter}
\end{figure}

A variety of robotic tasks, such as warehouse automation and service robotics, motivate computationally efficient and high-quality solutions to prehensile manipulation. Typical applications involve the need to pick and place a variety of objects from tabletop or shelf-like storage units. Such manipulation challenges are demonstrated in Fig. \ref{fig:table_clutter}. 

\begin{figure}[h]
    \centering
\vspace{0.1in}
    \includegraphics[width=0.42\textwidth]{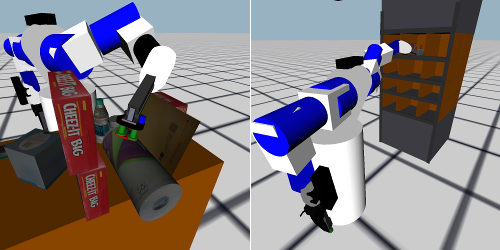}
    \includegraphics[width=0.42\textwidth]{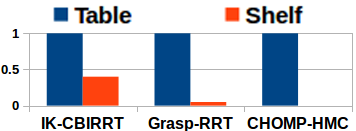}
    \caption{(top) Two manipulation challenges in a table and shelf environment. (bottom) Success rate (collision-free solution found within 30 seconds) of manipulation motion planning methods averaged over 50 trials.}
    \label{fig:intro_example}
\end{figure}

Consider a tabletop and shelf manipulation challenge, as
shown in Fig. \ref{fig:intro_example}(\textit{top}). Many existing methods
are capable of finding collision-free motions in such
tabletop challenges \cite{berenson2009manipulation}\cite{ratliff2009chomp}\cite{vahrenkamp2012simultaneous}. 
Prior work \cite{azizi2017geometric} has shown that large amounts of clutter can significantly reduce the
viable valid grasps on an object. 
While traditional strategies successfully find solutions in the tabletop setting, more cluttered environments like the 
shelf results in a massive degradation of performance, as highlighted in Fig. \ref{fig:intro_example}(\textit{bottom}). 
One possible explanation is that the clearance of the solutions found in cluttered scenes is low,
where clutter refers to the minimum distance between the robot geometries and the rest of the environment.
This work accordingly refers to scenes with low clearance as \emph{densely cluttered},
which have narrow passages in the workspace that make the motion planning problem harder. 
Nevertheless, coming up with high-quality motion planning solutions to such challenges
in a reasonable amount of time remains an important objective.


In problems where solutions exist in workspace clutter, there is a need to effectively guide the search process. 
The key insight is that in manipulation tasks, goals are typically  end-effector centric i.e., grasping configurations. Consequently, heuristic guidance in manipulation tasks should reason about the end-effector. 
Additionally, the planning algorithm has to be designed to effectively use such guidance to find high-quality solutions quickly.

Towards solving such a challenge, this work proposes the Jacobian Informed Search Tree (JIST) method, which is a
heuristic-guided sampling-based search algorithm for prehensile manipulation
planning in the presence of dense clutter. The key idea is to plan around the constraints
induced on the end-effector by the presence of clutter, so as to guide the 
motions of the arm towards areas which lead to a solution. The contributions of this work are: 
1) developing an effective heuristic for guiding the arm by solving the motion planning 
problem for just the end effector; 2) applying the pseudo-inverse Jacobian as
a steering process which allows the arm to be guided by the heuristic; and
3) incorporating task space guided maneuvers effective in clutter into an asymptotically optimal motion planner.

The performance of \jist\ is evaluated experimentally through comparison with
other state-of-the-art planners for picking and placing tasks,
including both trajectory optimization ({\tt{CHOMP}}-{\tt{HMC}}) and
sampling-based methods ({\tt {Grasp}}-{\tt{RRT}}, {\tt IK}-{\tt
CBiRRT}). The results indicate that \jist, in contrast to the
alternatives, can quickly and reliably compute high-quality solutions
in the presence of dense clutter in unknown scenes, for a variety
of robotic platforms, without requiring parameter tuning.

\section{Background and Relative Contribution}
\label{sec:background}
The proposed method draws inspiration from both broad categories of
approaches for planning arm motion: a) local optimization and b)
global search.

Local optimization follows a locally valid gradient towards finding a
solution. \emph{Artificial potential fields} incrementally move the
current robot configuration by following such a gradient towards the
goal.  For arm planning \cite{khatib1986real}, workspace virtual
forces are mapped to torques through the manipulator's Jacobian. The
framework allows for a hierarchy of objectives that a redundant arm
can potentially try to satisfy \cite{song2002potential}.  The typical
limitation is the presence of local minima, which can be avoided
through integration with a global planner \cite{warren1989global} or
defining a navigation function \cite{rimon1988exact}. The latter is
difficult to find for complex geometric
problems \cite{koren1991potential}. \jist\ uses gradient information
to locally guide the exploration of the arm's path.  This is performed
in the context of a global search process so as to achieve stronger
guarantees.

\emph{Trajectory optimization} methods define a gradient over the entire
trajectory, and are capable of finding smooth solutions for high DOF systems in sparse setups
\cite{ratliff2009chomp}\cite{schulman2013finding}\cite{dong2016motion}.
Such methods use precomputed signed distance fields for defining
obstacle avoidance gradients in the workspace.  The accuracy and cost
of computing these fields depend upon resolution. \jist\ similarly
aims for high-quality solutions but avoids dependency on parameters
and better handles clutter.

\emph{Global heuristic search} approaches aim to search the configuration
space of a robotic arm given a discretization, frequently by focusing
the search in the most promising subset or projection of that
space \cite{cohen2014single}\cite{gochev2014motion}\cite{cohen2015planning}. Heuristics that have been used in the context of
manipulation tasks include reachability \cite{hang2016evolution} or
geometric task-based reasoning \cite{garrett2015ffrob}.
\jist\ first solves the manipulation problem for a free-flying end
effector geometry, and then uses this as a heuristic during the search
in the arm's configuration space.

\emph{Sampling-based planners} \cite{Kavraki:1996aa}\cite{kuffner2000rrt} aim to scale better in high dimensions,
while providing guarantees, such as asymptotic
optimality \cite{karaman2011sampling}\cite{zak2018dirt}. They incrementally
explore the arm's configuration space through sampling. A variety of
sampling-based planners has been developed for robotic
arms \cite{simeon2004manipulation}\cite{stilman2010global}\cite{mcmahon2014sampling}, some of which use the pseudo-inverse of the
Jacobian matrix for steering \cite{vahrenkamp2012simultaneous} 
which has been shown to be a probabilistically
complete projection onto the grasping
manifold \cite{berenson2010probabilistically}. The current method
follows the sampling-based framework to maintain desirable guarantees,
while properly guiding the search to quickly acquire high-quality
solutions in terms of end-effector's displacement.

Some approaches deal with integrated grasp and motion planning \cite{fontanals2014integrated}\cite{haustein2017integrating}. This work focuses on motion planning aspects
and uses the output of grasping planners \cite{goldfeder2009columbia}\cite{berenson2008grasp}\cite{xue2011efficient}, i.e.,
grasps to set the goals of the corresponding picking challenges, which
are frequently defined as end effector poses relative to a target
object.

\section{Problem Setup}
\label{sec:problem}
Consider a robot arm with $d$ degrees of freedom, which must solve
pick and place tasks involving a target object $\obj$ in a workspace
with a set of static obstacles $\mathbb{O}$. 

\assumption{The geometric models of object $\obj$ and obstacles
 $\mathbb{O}$ are known. Online, before calling the planner, a
 perception module provides the 6D pose $p_o \in SE(3)$ of $\obj$, as
 well as the 6D poses of obstacles $\mathbb{O}$. \qed}

This setup can be easily extended to point cloud representations, permitting the utilization of recent work on 6D pose estimation \cite{mitash2018}. Note that this does not allow precomputation that reasons over the
object or obstacle poses, as this information is available only
online.

\subsection{Goal Task Space Motion Planning}

\definition[$C$-space]{An arm's configuration $q$ is a point in the
robot's $C$-space: $ \cfull \subset \mathbb{R}^d $.  The colliding
$C$-space ($\cobs \subset \cfull$) corresponds to configurations where
the arm collides with $\mathbb{O}$ or $o$ in their detected poses. The
free $C$-space is then defined as: $ \cfree = \cfull \setminus \cobs
$. \qed}

The motion planning problem to be solved in manipulation setups is
frequently not defined directly in $\cfree$. Instead, the task relates
to the arm's end effector pose.

\definition[Pose Space]{The end effector's pose space
$\mathbb{E} \subseteq \sethree$ corresponds to the reachable subset of
poses for the arm's end effector. \qed}

The pose space $\mathbb{E}$ can be computed given the forward
kinematics of the robotic arm $\fk: \cfull \rightarrow \mathbb{E}
$. The inverse kinematics function corresponds to the inverse mapping:
$\ik: \mathbb{E} \rightarrow \cfull$. For redundant manipulators, $
dim(\mathbb{E}) < dim(\cspace)$, so $ \ik $ describes a mapping to
self-motion manifolds of the
manipulator \cite{berenson2010probabilistically}, composed of states,
which all have the same end effector pose.

\definition[Goal-Constrained Task Space (GCTS)]{Given a set of goal poses
$\mathbb{E}_{goal} \subset \mathbb{E}$ for the end effector, the arm's
goal task space $\tspace_{goal} \subset \cfull$ denotes the robot
configurations that bring the end effector to a pose in the set
$\mathbb{E}_{goal}$: $\tspace_{goal} = \{ \forall\ q \in \cfull\
|\ \fk(q) \in \mathbb{E}_{goal} \}.$ \qed}

For a redundant arm, even a discrete pose set $\mathbb{E}_{goal}$
gives rise to a continuous submanifold of solutions
$\tspace_{goal}$. Given the above definition, the objective is to solve the
following:

\definition[Motion Planning with a GCTS]{Given a start
arm configuration $\cstart \in \cfree$ and a set of goal poses for the
end effector $\mathbb{E}_{goal} \subset \mathbb{E}$, a solution path
is a continuous curve $\pi: [0,1] \rightarrow \cfree$, so that
$ \pi(0) = \cstart $ and $ \pi(1) \in \tspace_{goal}.$ Given the space
of all solution paths $\Pi$, the objective is to find the path $\pi^*
= \argmin_{\pi \in \Pi} C(\pi)$, which minimizes a cost function $C$
defined over $\Pi$. \qed}



\subsection{Defining Goal Poses for Pick and Place Instances}

For picking, the goal is to move the end effector to a pose that
allows for a stable grasp.

\assumption{A grasping planner produces a set of stable, collision-free
grasps for the arm's end effector and the target object $o$ at the
object's detected pose $p_o$. \qed}

\begin{figure}[h]
\centering
\includegraphics[width=0.37\textwidth]{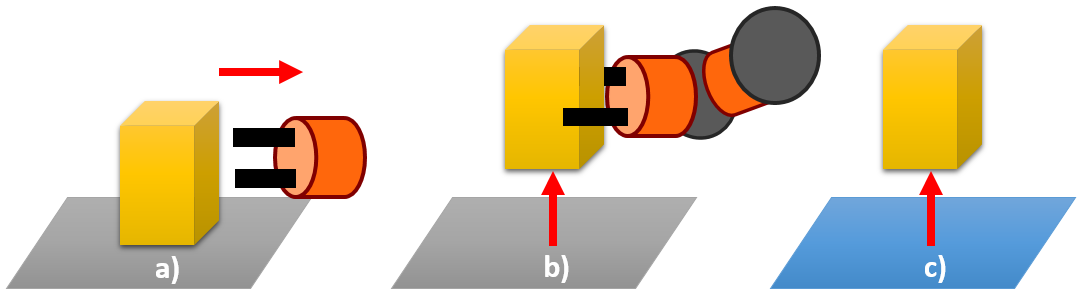}
\caption{a) Pre-grasp pose, b) post-grasp state, c)
object pre-placement.}
\label{fig:positions}
\end{figure}

A database of grasps can be computed in the object's
frame \cite{miller2003automatic}\cite{goldfeder2009columbia}\cite{berenson2008grasp}\cite{xue2011efficient} and evaluated in terms of their
stability offline \cite{liu2015fast}. Online, the grasps
are transformed given the object pose $p_o$ and collision checked,
given the end effector's geometry and the obstacles
$\mathbb{O}$. Nevertheless, states that correspond to grasps are on
the boundary of $\cfree$ with $\cobs$, which result in very hard
motion planning problems. For this reason, the goal set for picking
instances is defined for ``pre-grasp'' poses: each grasp returned by
the grasping planner is transformed by moving the end effector along a
vector away from the object as shown in Fig. \ref{fig:positions}. A
solution returned by the planner that brings the end effector to a
``pre-grasp'' pose, needs to be appended with a motion so that the end
effector achieves the actual grasp.

A similar setup is used for placing, where in the initial state the
target object is immobilized by the end effector. In this case, the
arm-object system can be considered as the robot and the end effector
is the object itself.

\assumption{For placement, a task planner produces a set of
stable object placements on a support surface. \qed}

Again, it is helpful to avoid defining problems that bring the robot
close to the boundary of $\cfree$. Thus, if the initial state has the
object on a resting surface, the state is transformed so that the
object is raised away from the surface. For a goal object placement,
the pose is transformed so that the object is again raised away from
the surface. Planning solutions from ``post-grasp'' states to
``pre-placement'' goal poses are again extended so that the robot
moves the object from the grasp state to the actual goal placement.
 
\subsection{Distance and Cost Function}


It has been established that the problem at hand is closely tied to the motions of the end effector, since the goals and heuristic can be described by it. The current work proceeds to define a distance function appropriate for this setting, that minimizes end effector displacement. Another benefit of this is that paths that minimize
end effector motion tend to look more natural to human observers who pay more attention to the end effector rather than the rest of
the arm \cite{zhao2016experimental}.


Towards this objective,
this work employs the $\cdist$ distance function, which is a
model-aware distance metric for $SE(3)$, which can be approximated
efficiently by the $\mathtt{C-DIST}$ algorithm \cite{zhang2007c}.
Given the convex-hull $H$ of the end-effector's model,
the $\cdist$ distance $D$ of two poses $e_i, e_j \in SE(3)$ is:
\begin{equation}
\label{equ:cdist}
D(e_i,e_j) = max_{{\bf p} \in H}||{\bf p}(e_i) - {\bf p}(e_j)||_2
\end{equation}

With slight abuse of notation, the
expression $D(q,e)$ between $q \in \cfull$ and end effector pose $e \in
\mathbb{E}$ will denote the distance $D(\fk(q),e)$. It follows that
the distance between $q_i, q_j \in \cfull$ is $D(\fk(q_i),\fk(q_j))$ Then, the cost of
a path $\pi$ is:
\begin{equation}
\label{equ:cost}
C(\pi) = \sum_{i=0}^{k}D(\pi(i),\pi(i+1))
\end{equation}

where $k$ is a discretization along $\pi$. For the remainder
of the paper, unless otherwise stated, \emph{distances} will refer to Eq. \ref{equ:cdist}, and 
\emph{costs} will refer to Eq. \ref{equ:cost}.

\section{Jacobian Informed Search Tree}
\label{sec:method}
This section describes a process for exploring the end effector's
space $\mathbb{E}$ and how the corresponding information is used to
guide the exploration of the arm's space \cfull.

\subsection{Exploring the End Effector's Pose Space}

A simple way to estimate how far the end effector is at pose
$e \in \mathbb{E}$ from the set $\mathbb{E}_{goal}$ is to consider the
minimum SE(3) distance from $e$ to all goal poses
$\mathbb{E}_{goal}$. Nevertheless, in the presence of clutter, these
distances are not informative of the end effector's path cost so as to
reach goal poses.

\begin{figure}[h]
\centering
\includegraphics[width=0.16\textwidth]{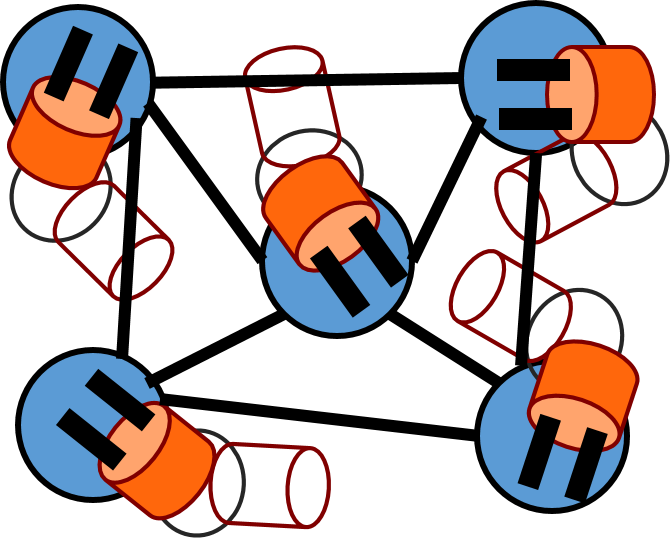}
\includegraphics[width=0.31\textwidth]{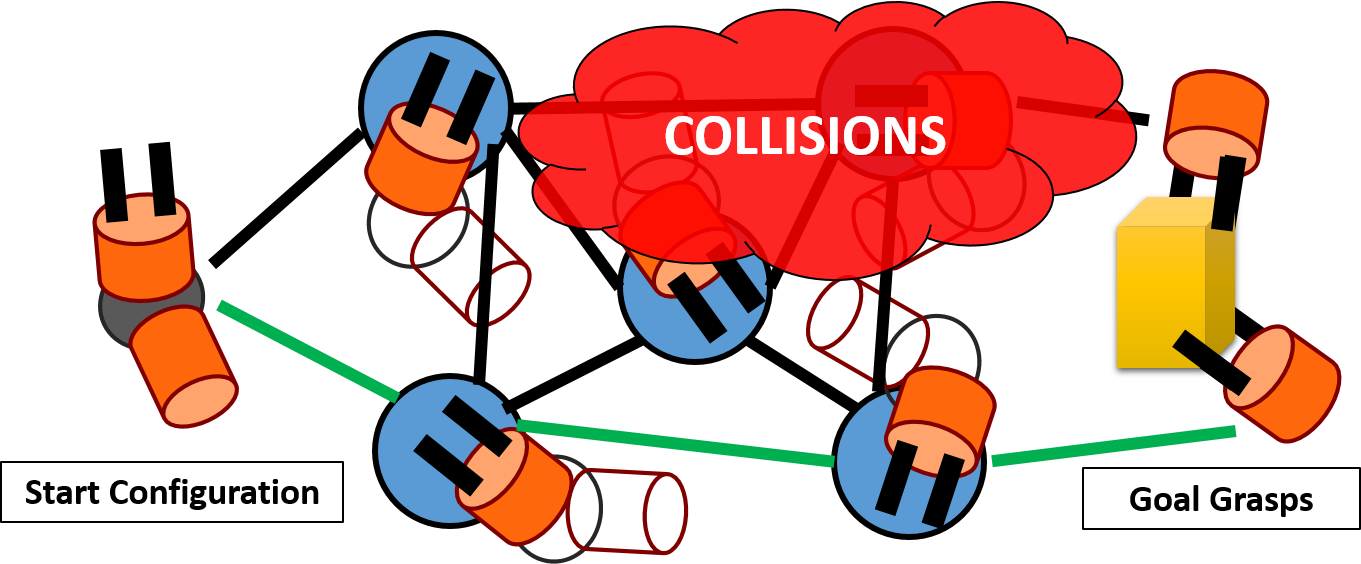}
\caption{(left) Construction of $\eeroadmap$. (right) Online search
of $\eeroadmap$.}
\label{fig:ee_roadmap}
\end{figure}

To take clutter into account, this work proposes the offline
construction and online search of a roadmap $ \eeroadmap( V_{ee},
E_{ee} )$ in $\mathbb{E}$ as shown in Fig. \ref{fig:ee_roadmap}. The
vertices $V_{ee}$ store reachable end effector poses given the arm's
kinematics, while the edges store local connection paths for the
free-flying end effector. The goal poses $\mathbb{E}_{goal}$ and the
start pose $e_{start} = \fk(\cstart)$ are attached to $\eeroadmap$,
which is then searched online for a path from $\mathbb{E}_{goal}$ to
$e_{start}$. Performing the search in this direction produces a
multi-start search tree rooted at $\mathbb{E}_{goal}$. During the
online search, only collisions between the end effector, the obstacles
$\mathbb{O}$, and the object $o$, but not the arm, are taken into
account.  This allows for quick estimations of the costs to reach
$\mathbb{E}_{goal}$. These costs are used later as heuristics to
guide a search procedure in the arm's $C$-space.

\textbf{Offline Construction of $\eeroadmap$:} 
The end effector reachability roadmap $\eeroadmap( V_{ee}, E_{ee} )$
is constructed offline without knowledge of the workspace and reflects
the robot's reachability. A sampling-based process is followed similar
to {\tt PRM}$^*$ \cite{karaman2011sampling}. A random arm state
$q_{r} \in \cfull$ is sampled and the corresponding end effector
pose $\fk(q_{r}) \in \mathbb{E}$ is stored as a vertex. Each
vertex is then connected with an edge to the closest $log|V_{ee}|$
vertices in terms of $D(q_{r},e)$. Each edge stores an
interpolation in SE(3) between the two vertex poses according to a
discretization defined by an estimation of the end effector's maximum
velocity.

\textbf{Online Computation of Distances to Goal Poses:} 
Given knowledge of the workspace, $\eeroadmap$ is searched for
collision-free end effector paths from the goal poses
$\mathbb{E}_{goal}$ to the start pose $e_{start}$. The poses
$\mathbb{E}_{goal}$ and $e_{start}$ are added as vertices on
$\eeroadmap$ and are connected to their closest $log|V_{ee}|$ neighbors.

Then, a multi-start, multi-objective $\astar$ ($ \mathtt{MSMO\_\astar}
$) search procedure is performed on $\eeroadmap$. Upon initialization,
the priority queue of the search includes all goal poses
$\mathbb{E}_{goal}$. Then, each search node corresponds to a path from
a pose in $\mathbb{E}_{goal}$ to a vertex $u \in V_{ee}$. The priority
queue sorts nodes so as to minimize the following two ordered
objectives:
\begin{myitem}
\item $f_1(u)$, the number of colliding  poses along the
path from $\mathbb{E}_{goal}$ to $u$, given the roadmap's edge
discretization;
\item $f_2(u) = g_2(u) + h_2(u)$, where $g_2$ is $C(\pi(\mathbb{E}_{goal},e_u))$ and $h_2$ is $D(e_u, e_{start})$.
\end{myitem}
Once a vertex $u$ has been expanded, it is added to the ``closed
list'' $\mathbb{L}_{closed}$. If there is a collision-free end effector path from
$\mathbb{E}_{goal}$ to $e_{start}$, the above search will report the
shortest path. Then, the $g_2$ cost of
vertices $u$ in $\mathbb{L}_{closed}$ corresponds to an estimate of
the distance along the shortest collision free path from $u$ to
$\mathbb{E}_{goal}$, i.e., $g_2$ is a discrete approximation of a
navigation function in $\mathbb{E}$ given the presence of
obstacles. The proposed approach uses these $g_2$ values to
heuristically guide the exploration in $\cfull$. In dense clutter, if the roadmap's resolution does not permit
collision-free paths, the algorithm returns the path with the minimum
number of collisions given its discretization. In this case, the $g_2$
values of vertices in $\mathbb{L}_{closed}$ will guide the arm
exploration along the shortest, least colliding solutions for the end
effector.

\subsection{Generating Arm Paths}

\begin{figure*}
\centering
\vspace{0.1in}
\includegraphics[width=0.8\textwidth]{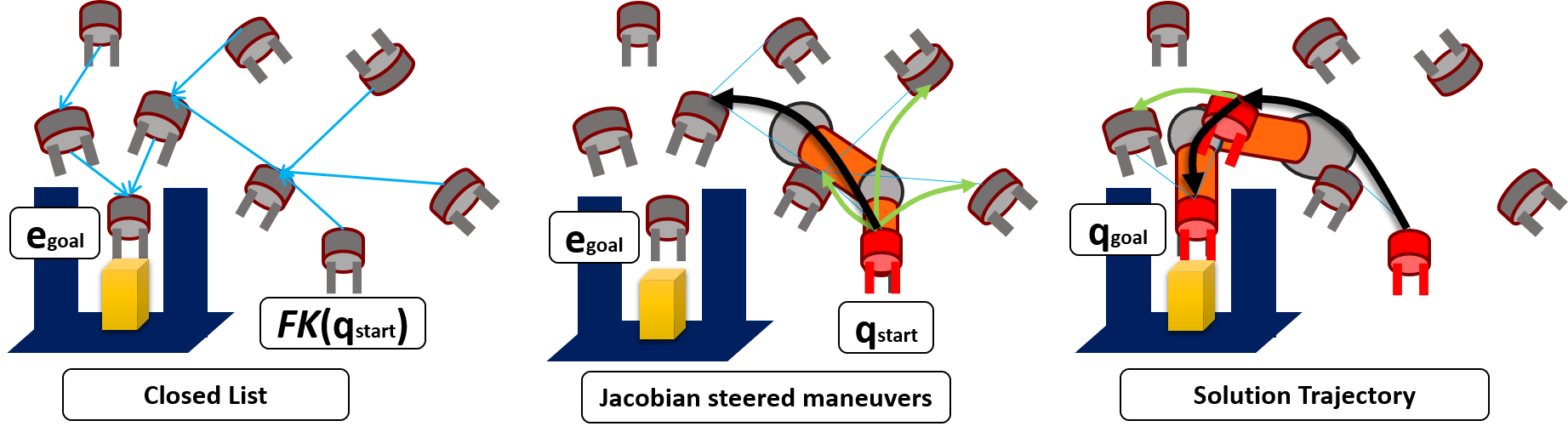}
\caption{An illustration of the \jist\ framework for solving
manipulation queries: (left) the closed list $\mathbb{L}_{closed}$ is
constructed using a multi-start, multi-objective A* on $\eeroadmap$
starting from $\mathbb{E}_{goal}$ to $\fk(q_{start})$; (middle) \jist\
expands arm paths from the start configuration $q_{start}$ guided by
the cost-to-go stored in $\mathbb{L}_{closed}$; (right) the best path
found $\pi^*$ is kept track of as the solution trajectory, with
subsequent solutions from \jist\ improving upon it. }
\label{fig:jist_tree}
\end{figure*}

This section describes how to generate paths for the arm given the
vertices $u \in V_{ee}$ stored in the ``closed list''
$\mathbb{L}_{closed}$ and the cost estimates $g_2$ from $u$ to
$\mathbb{E}_{goal}$. An illustration of the steps of the algorithm can
be seen in Figure \ref{fig:jist_tree}.

\begin{algorithm}[]
\caption{$\jist$ ($q_{start}, \mathbb{E}_{goal}, R_{ee}, N, \kappa $)}
\label{algo:jist}
$ \mathbb{L}_{closed} \leftarrow {\tt MSMO\_A^*}( R_{ee},
q_{start}, \mathbb{E}_{goal}) $\;
$ \pi^*\leftarrow \emptyset $\;
$ \mathtt{T}\leftarrow\{q_{start}\};\ q_{new}\leftarrow q_{start} $\;
$ \acand(q_{new}) \leftarrow NULL $\;	
\For{$ N $ iterations}
{
\If{$ q_{new}\neq\emptyset\ \band\ \hval(q_{new}, \mathbb{L}_{closed}
)<\hval(parent(q_{new}), \mathbb{L}_{closed} ) $}
{
	$ q_{sel}\leftarrow q_{new} $\;
}
\Else
{
	$ q_{sel}\leftarrow {\mathtt{SearchSelection()}} $\;
}
\If{$ \acand(q_{sel}) = NULL $}
{
	$ \acand(q_{sel})\leftarrow {\mathtt{GreedyEdges}}(q_{sel}, \mathbb{L}_{closed}, \kappa ) $\;
}
\ElseIf{$ \acand(q_{sel}) = \emptyset $}
{
	$ \acand(q_{sel})\leftarrow {\mathtt{FallbackEdges}}(q_{sel}) $\;
}
{
	$ a_{best} \leftarrow\underset{a\in\acand}{argmin}  \ \hval(a, \mathbb{L}_{closed})$\;
	$ \acand(q_{sel}) \leftarrow \acand(q_{sel}) \setminus a_{best} $\;
	$ q_{new}\leftarrow {\mathtt{Steer}}(q_{sel},a_{best}) $\;
        \If{${\mathtt{BaB}}(q_{new}, \pi^*)\ \bor\ {\mathtt{CC}}(q_{sel}	\rightarrow q_{new}$)}
	{
		$ q_{new}\leftarrow\emptyset $\;
	}
	\If {$ q_{new}\neq\emptyset$}
        {
		$ \pi^*\leftarrow{\mathtt{AddEdge}}(\mathtt{T},q_{sel}
	\rightarrow q_{new}) $\;
		$ \acand(q_{new}) \leftarrow NULL $\;	
        }
}
}
{\bf return} $\pi^*$\;
\end{algorithm}

The algorithm is outlined in Alg. \ref{algo:jist}, which
is inspired by a search strategy \cite{zak2018dirt} capable of 
utilizing heuristic guidance. It receives as
input the start arm configuration $\cstart$, the goal poses
$\mathbb{E}_{goal}$, the end effector reachability roadmap $R_{ee}$,
the number of iterations $N$, and the parameter of the maximum number
$ \kappa $ of greedy edge expansions per iteration. The first step of
the method is to execute the multi-start, multi-objective $A^*$ search
over $R_{ee}$ outlined in the previous section, in order to acquire
the closed list $\mathbb{L}_{closed}$. This contains the vertices of
$R_{ee}$ visited during the search and the corresponding cost
estimates to the goal poses $\mathbb{E}_{goal}$. The method then
builds a search tree whose nodes correspond to arm configurations. The
tree is rooted at the start configuration $ \cstart $ (line 3) and is
expanded towards configurations that bring the end effector to goal
poses.

Each iteration of the algorithm starts by selecting a node of the
search tree so as to expand it (lines 6-9). If a node was added during
the previous iteration, and its heuristic value is better than its
parent on the tree, then the newly added node is selected for
expansion (line 6-7). The heuristic value
$\hval(q, \mathbb{L}_{closed})$ of a configuration $q$ corresponds to
an estimate of the cost to reach $\mathbb{E}_{goal}$
approximated using $\mathbb{L}_{closed}$.

If there was no progress made in the previous iteration towards
reaching $\mathbb{E}_{goal}$, the $ \mathtt{SearchSelection} $
subroutine (line 9) instead uses a probabilistic selection process,
where the probability of selecting a node depends on the node's
corresponding sum of cost from the root and heuristic cost-to-go.  All
nodes are guaranteed to have a non-zero probability of being selected;
however, nodes with better costs and heuristic sum have a greater
probability.

Once a configuration $q_{sel}$ is selected for expansion, the set
$ \acand $ represents a set of actions that can be used to extend the
tree out of $q_{sel}$, which can correspond either to target poses for
the end effector or joint velocities for the arm. The first
time that node $q_{sel}$ is selected for expansion (line 10), the
subroutine $ \mathtt{GreedyEdges}$ is used to generate target poses
for extending the tree out of $q_{sel}$ (line 11). 

\jist\ iterates over the
generated actions, which are prioritized in terms of the lowest
$ \hval $ in terms of the resulting pose and the best one $ a_{best} $
is considered for addition at each iteration (line
14). $ \mathtt{Steer} $ uses an appropriate steering method to
generate a trajectory where the state reached at the end corresponds
to $ q_{new} $ (line 16). The configuration $q_{new}$ must satisfy two
conditions so as to be added as a new node (line 17): (a) A branch and
bound process ($\mathtt{BaB}$) ensures that $q_{new}$ has smaller path
cost relative to the length of the best solution found $\pi^*$; (b)
A collision checker $ \mathtt{CC} $ verifies whether the
path from $q_{sel}$ to $q_{new}$ is in collision. If the node
$q_{new}$ passes these checks (line 19), then it is added to the tree
$ \mathtt{T} $. If a better path to the goal is discovered with the
addition of $q_{new}$, it is recorded as $ \pi^* $ (line 20).


			

\textbf{Greedy Edge Generation:} 
These greedy edges are guided by the information stored in the
$ \mathbb{L}_{closed} $, which contains end effector
poses explored during the online search of $\eeroadmap$. First, the
nearest end effector pose $e_{near} \in \mathbb{L}_{closed}$ is found
relative to $FK(q_{sel})$. The method obtains poses to steer towards by examining the 
adjacent vertices of $e_{near}$ on $R_{ee}$, that belong to $ \mathbb{L}_{closed} $ and also have a better heuristic estimate
 than $q_{sel}$. 

The number of target poses added is controlled by parameter
$\kappa$. Experimental indications show that a value $\kappa=2*log|V_{ee}|$
works well in practice. The method then updates $e_{near}$ to its predecessor from the closed set, 
and attempts to add target poses until either $\kappa$ poses have been discovered, or there are no
 other predecessors available. In the latter case, the remaining target poses are sampled
directly from the set $\mathbb{E}_{goal}$.  Overall, this is a greedy
procedure, which traces along the search tree produced by the A*, while
examining nearby poses, which have also been expanded during the A*
search.

\textbf{Fallback Edge Generation:} Every time a node is selected and all
greedy edges out of it have already been considered, the method
reverts to considering two fallback strategies for generating edges.
The first strategy corresponds to a random control in terms
of joint velocities executed for a random duration. The second
strategy randomly selects a goal
state $q_{goal} \in \mathbb{Q}_{goal}$ already discovered by the
algorithm as a target to steer towards.

\textbf{Steering:} For the fallback edges, the steering subroutine
either uses $C$-space interpolation to goal arm states, or executes
the random controls.

The greedy edges consist of target end effector poses. The
approach uses a steering method based on the pseudo-inverse of the
manipulator Jacobian matrix to achieve the target poses. Given an arm
configuration $q = (q_1, ... , q_d)^T$ and the corresponding end
effector pose $e = FK(q)$, the Jacobian matrix is
$J(q) = \frac{\partial e}{\partial q}$.

For a target end effector pose $e_{target}$, let $\Delta e =
 e_{target} - e$ denote the desired change in position of the end
 effector.  The objective of the Jacobian-based steering
 process \jacsteering\ is to compute the joint controls which solve
 $ \Delta e = J \Delta q$. The method uses the pseudo-inverse of the
 Jacobian, $\pseudoj$, so as to minimize $|| J \Delta q - \Delta e
 ||^2$, where $\Delta q = \pseudoj \Delta e$. To account for
 singularities, damping \cite{buss2005selectively} and clamping are
 also employed. Thus, at time $ t $, for the configuration $ q_t $ and
 the corresponding end effector pose $ e_t $, the control update rule
 for \jacsteering\ is:
\begin{equation}
\label{equ:jacsteering}
\Delta q_t = \mathbb{J}^+(q_t) \cdot (e_{target} - e_t)
\end{equation}


Since the objective is to minimize \cdist, the gradient followed by
the above rule reduces the distance of the end effector from the the
target pose. Any configuration $ q_{goal} \in \mathbb{Q}_{goal}$ discovered through
\jacsteering\ is kept track of. A benefit to using $\jacsteering$ is that it satisfies
the necessary properties of the projection
operator \cite{berenson2010probabilistically} needed to ensure
coverage of $\mathbb{Q}_{goal}$. $\jacsteering$\ is used in the
connection of ``pre-grasps'', ``post-grasps'', and ``pre-placemements'' as
described in Section \ref{sec:problem}.

\section{Experiments}
\label{sec:experiments}
\begin{figure*}[t]
\centering
\includegraphics[width=0.98\textwidth]{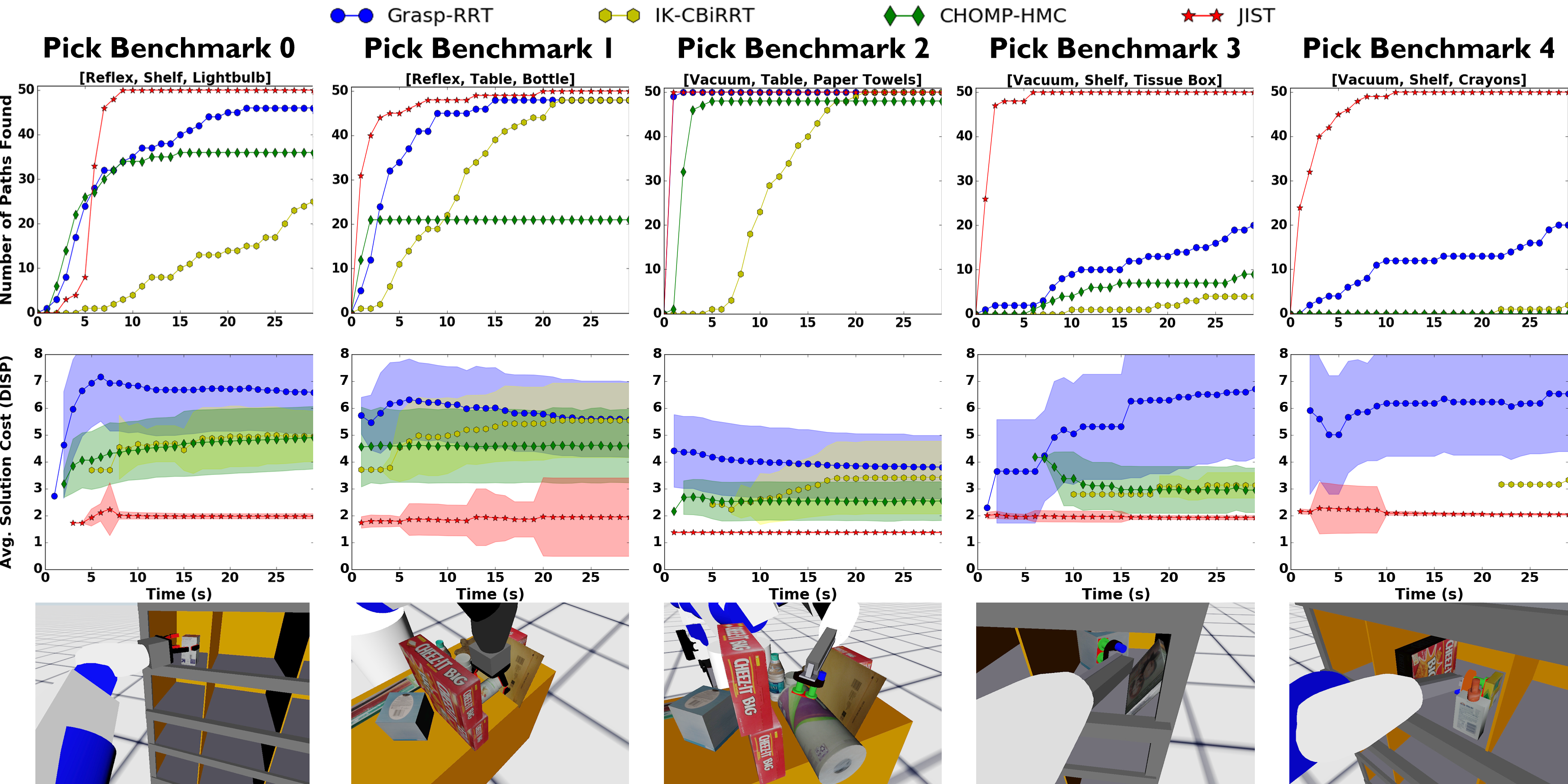}
\caption{Success Rates, Solution Costs, and Example Picks for the Kuka+ReFlex (0), Baxter+Parallel (1), Motoman+ReFlex (2), Motoman+Vacuum (3,4)}
\label{fig:pick_results}
\end{figure*}

\jist\ is compared against other planners for
solving manipulation challenges. Recent work \cite{haustein2017integrating} has
utilized {\tt CBiRRT} \cite{berenson2009manipulation} for solving grasping problems. 
Accordingly, this work compares against an IK-based variant of {\tt CBiRRT}, which uses IK 
on the $\mathbb{E}_{goal}$ to
construct roots of the goal tree.  {\tt {Grasp}}-{\tt{RRT}} corresponds to an
{\tt RRT} variant, which utilizes $ \jacsteering $ to discover grasp
states during the goal biasing
phase \cite{vahrenkamp2012simultaneous}. {\tt{CHOMP}}-{\tt{HMC}}
corresponds to an OpenRave \cite{diankov_thesis} prob. complete
implementation of \chomp\cite{ratliff2009chomp}.

\textbf{Setup:} A common planning
software was used for the sampling-based
planners \cite{littlefield2014extensible}. All methods were evaluated
on a single Intel Xeon E5-4650 processor with 8 GB of RAM. Experiments
were conducted on a variety of robotic manipulators: Kuka LBR iiwaa (7 DOF), Rethink Baxter (14 DOF),
and Motoman SDA10F (16 DOF). Three types of end effectors were
evaluated: the ReFlex hand, a parallel-jaw, and a vacuum suction-cup.  Each
benchmark involved computing a trajectory for one of the arms to a set
of goal end effector poses. For pick benchmarks, the poses correspond
to pre-grasps 2 cm away from the actual grasps.
For place benchmarks, the goal poses corresponded to pre-placements
for the object 5 cm above the resting surface.

\textbf{Metrics:} The number of planning successes and solution quality
were measured over time, and an average over 50 runs was reported at
each time interval. A success was counted if the planner computed a
collision-free trajectory from the start state to the goal
end-effector pose within the corresponding time.  Solution quality was
measured using \cdist\ .

\textbf{Initialization:} Both \jist\ and \chomp\ required some
additional initialization prior to attempting to solve any of the
benchmarks. \jist\ required $R_{ee}$ to be first computed offline
(without knowledge of the scene). For the experiments, the size of
$R_{ee}$ was $ 25,000 $ vertices, and the maximum number of maneuvers
$\kappa$ was set to 20. For \chomp\, each environment had a
signed-distance field constructed for it, and also had its parameters
tuned per benchmark, since no single set of parameters sufficed for
all.  For \chomp\ and {\tt IK}-{\tt CBiRRT}, reachable, and
collision-free arm configurations were computed through IK at goal
poses.

\begin{figure}[h]
\centering
\includegraphics[width=0.23\textwidth]{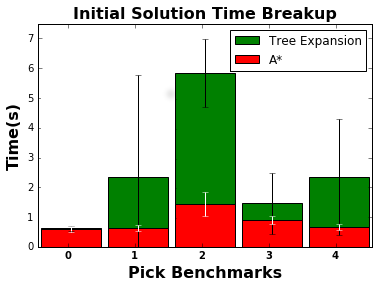}
\includegraphics[width=0.23\textwidth]{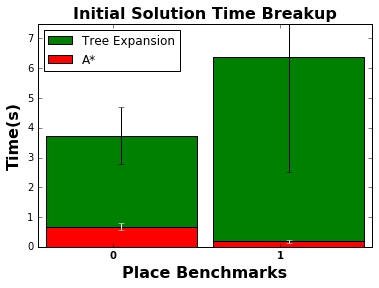}
\caption{Breakdown in time for the initial solution using \jist\ for pick (left) and place (right) benchmarks.} 
\label{fig:average_solution_time}
\end{figure}

\textbf{Pick Benchmarks:} The results for success rate and solution
cost over time are shown in Figure \ref{fig:pick_results}.  Across all
experiments, \jist\ converged to a 100\% success rate within the time
limit of 30 seconds, with an average initial solution time breakdown
shown in Figure \ref{fig:average_solution_time} (left), while also
providing the lowest cost solutions in all cases.  The end effector
heuristic was shown to be effective in guiding the search to produce
high quality initial solutions. An interesting side-note here regarding
\chomp\ is that when initialized with solution states found
by \jist\ and with some additional parameter tuning, the performance of \chomp\ improved significantly. 
This indicates that solutions out of \jist\ can serve
as better initializations for trajectory optimization methods.


\begin{figure}[t!]
    \centering
    \includegraphics[width=0.48\textwidth]{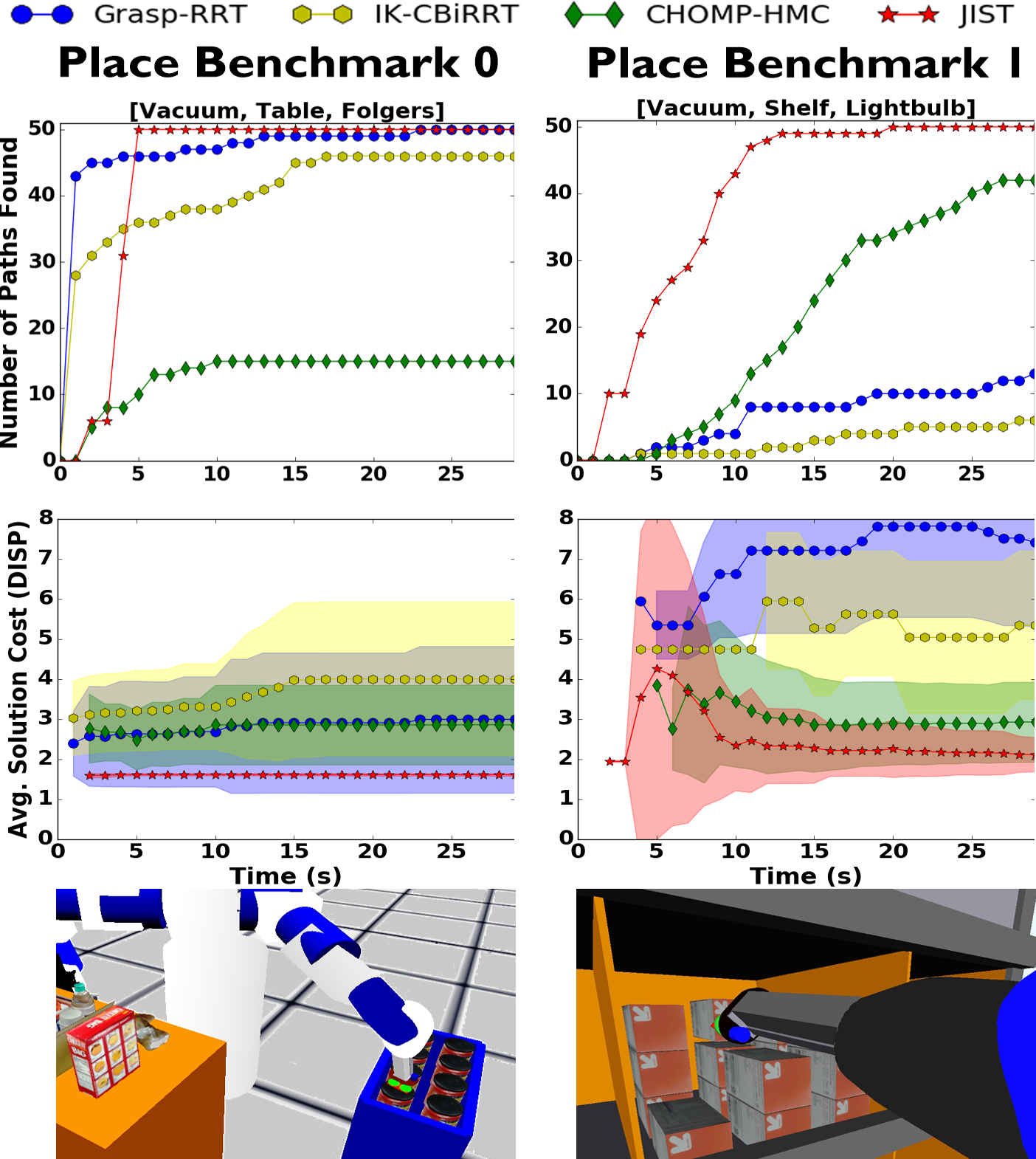}
    \caption{Success Rates, Solution Costs, and Example Placements for the Vacuum Place Benchmarks in the Table (0) and the Shelf (1) environments.} 
    \label{fig:results_pnp}
\end{figure}

\textbf{Place Benchmarks:} The results for the place benchmarks are shown in Figure \ref{fig:results_pnp}, with average solution times shown in Figure \ref{fig:average_solution_time}.
On average, \jist\ spent longer in the tree expansion portion of the
method, relative to the pick benchmarks due to the attached target
object causing more collisions. Despite this initial slowdown, \jist\
still managed to converge to a 100\% success rate before all other
methods, while also still providing the lowest cost solutions.

\textbf{Remarks:} The average clearance (i.e. shortest distance between
the robot and the obstacles) of solutions found by all methods in the table environment
was 5.7 cm, whereas in the shelf environment it was 2.2 cm. Accordingly, the table
benchmarks were the easiest for all methods to solve. This is primarily due to the
fact that overhand grasps over the table were viable solutions. However, in the shelf benchmarks,
\jist\ maintains its high performance despite the clutter severely reducing the overall clearance
and viable grasps.

\section{Discussion}
\label{sec:discussion}
This paper presented the Jacobian Informed Search Tree (\jist)
algorithm, which is an asymptotically optimal, informed sampling-based
planner for controlling an arm in densely cluttered scenes so as to
achieve desired end effector configurations.  The method optimizes a
cost function representing the displacement of the end-effector, and
uses a heuristic computed by effectively searching the end effector's
task space.  \jist\ employs Jacobian-based steering to bias the
expansion of the tree towards end effector poses that appear promising
given the heuristic guidance.  The method was shown to have high
success rate, even in densely cluttered scenes, with fast initial
solution times and high quality solutions.

There are several directions for further exploration. For instance,
goal constraints can be introduced for different arm links, such as a
camera attached to the arm. One of the benefits of using
Jacobian-based steering is that it allows for the satisfaction of
secondary objectives in the null-space of the primary objective to
reach the target pose. In this way, \jist\ could also incorporate
additional constraints during task execution, such as maintaining
certain orientations of the grasped object or confining the
end-effector to a particular workspace region. Recent work in more
complex IK-solvers could also be utilized to more efficiently
guide the search tree \cite{Rakita-RSS-18}. Another direction is the
consideration of alternative cost functions, especially time-based and
multi-objective functions. Finally, it is interesting to evaluate the
performance of \jist\ without knowledge of the obstacle geometries by
operating directly over point clouds \cite{kuntzfast}.

%







{\small
\bibliographystyle{IEEEtran}
\bibliography{gmp}}
\end{document}